\documentclass[letterpaper, 10 pt, conference]{ieeeconf}  
\usepackage[T1]{fontenc}
\usepackage{lmodern}
\usepackage{graphicx}
\usepackage{booktabs}
\usepackage{color}
\usepackage{amsmath,amssymb}
\usepackage{dsfont}
\usepackage{subcaption}
\usepackage{multirow}
\usepackage{makecell}
\usepackage{url}

\IEEEoverridecommandlockouts                              

\title{Learning from Mixed-Quality Deployment Experience for Robot Manipulation}

\author{Yangang Ren$^{1}$, Yujie Yan$^{1}$, Zirui Li$^{1}$, Jiaming Guo$^{1}$, Di Zeng$^{2}$, Ji Tao$^{2}$, Lan Yu$^{3}$, Xuesong Tian$^{3}$, Chen Lv$^{1,\dagger}$
\thanks{$^{1}$School of Mechanical and Aerospace Engineering, Nanyang Technological University, Singapore 639798.}
\thanks{$^{2}$Chongqing Changan Automobile Co., Ltd, Chongqing 400023, China.}
\thanks{$^{3}$Guangzhou Cloudbutterfly Technology Co., Ltd., Guangzhou 510220, China.}
\thanks{$^{\dagger}$Corresponding author: {\tt\small lyuchen@ntu.edu.sg}}
}

\begin{document}

\maketitle
\thispagestyle{empty}
\pagestyle{empty}

\begin{abstract}

Robot policies deployed in real environments naturally accumulate mixed-quality experience, including successful executions, partial progress, and failures. Although these rollouts provide valuable information for further learning, directly incorporating them into imitation learning may reinforce undesirable behaviors, while offline reinforcement learning often suffers from unreliable value estimation under sparse rewards and limited data coverage. We consider a practical post-deployment setting where learning relies only on naturally accumulated autonomous rollouts, without additional human corrections or exploratory interaction. To effectively exploit such experience, we propose Predictive Action Chunk Learning (PACL). PACL first learns a predictive chunk-level critic that evaluates temporally extended action sequences and augments temporal difference learning with future latent prediction, providing richer supervision for long-horizon value estimation. The learned critic then converts chunk-level Q-values into discrete quality conditions, which guide a diffusion actor to learn jointly from these mixed-quality experiences without treating all behaviors as equivalent supervision. At inference, the actor generates multiple action chunks and the critic selects the highest valued candidate. Experiments across simulated and real-world robot manipulation tasks show that PACL consistently improves the pretrained policy and outperforms strong imitation learning and offline reinforcement learning baselines.

\end{abstract}

\section{INTRODUCTION}

Recent advances in robot learning have produced increasingly capable policies across a broad range of embodied tasks \cite{chi2025diffusion,brohan2023rt2}. Many of these are built upon imitation learning from large collections of human demonstrations, including generalist policies trained on diverse multi-task and multi-robot data \cite{ghosh2024octo,kim2025openvla,black2025pi0}. While such demonstrations provide a strong initialization, they are costly to collect and cannot cover the full range of operating conditions, disturbances, and execution uncertainties encountered during actual robot operation. Once deployed, robots continuously accumulate new experience through their own executions, ranging from successful trials and partial progress to failures. Learning continually from experience acquired during operation has long been recognized as an important capability for autonomous robots \cite{lesort2020continual}. Reusing naturally accumulated deployment data therefore provides a practical path toward further policy improvement.

Yet deployment experience is inherently mixed in quality and cannot be directly incorporated into imitation learning. Using only successful rollouts is often insufficient, since they largely reproduce behaviors that the current policy already performs well and may provide limited additional supervision \cite{mirchandani2025scale,yue2024diverse}. Filtering imperfect actions can further reduce harmful imitation, but it also discards failure-inducing segments that reveal which behaviors should be avoided \cite{wu2025learning,chen2025s2i}. Offline reinforcement learning (RL) offers a natural alternative by retaining both successful and failed experience and propagating their outcomes through temporal difference learning. However, for visuomotor policies, sparse task rewards and limited deployment experience coverage make it difficult to reliably assign delayed outcomes to the action sequences that caused them \cite{huang2025rise}. Indeed, systematic evaluations on robot manipulation have found that representative offline RL methods often underperform strong imitation learning baselines on multi-source collected datasets \cite{mandlekar2022what}. This leaves a substantial gap between collecting mixed-quality deployment experience and reliably turning it into better robot policies.

These limitations have motivated post-training strategies that seek more informative supervision during deployment rather than relying solely on fixed offline data. One line of work uses human guidance, including expert interventions, recovery demonstrations, and targeted corrections, to provide explicit signals for policy refinement \cite{spencer2020learning,hu2026rac,physicalintelligence2025pi}. Another line optimizes pretrained policies through newly collected on-policy rollouts and real-world RL, often requiring repeated interaction, reward feedback, and environment resets \cite{tan2025interactive,lei2025rl100}. While effective, these methods depend on human supervision or purpose-driven physical interaction in the operational environment. In contrast, we learn directly from experience naturally accumulated during ordinary autonomous deployment, without deliberately collecting corrections or exploratory rollouts. This setting turns routine robot operation into a source of data for autonomous post-deployment improvement.

In this paper, we propose Predictive Action Chunk Learning (PACL), a post-deployment offline reinforcement learning method for learning from mixed-quality robot experience. PACL extends implicit Q-learning \cite{kostrikov2022offline} to temporally extended action chunks, enabling the critic to evaluate the action sequences generated by diffusion policies. To improve value learning under sparse rewards, the critic additionally predicts the future latent change induced by each action chunk, providing dense visual supervision of its consequences. The learned Q-values are then converted into discrete quality conditions that guide a diffusion actor to learn from human demonstrations and mixed-quality deployment rollouts without treating all behaviors as equivalent supervision. At inference, the conditioned actor proposes multiple action chunks and the predictive critic selects the highest-valued candidate. Our main contributions are threefold:

1) We propose Predictive Action Chunk Learning (PACL), which combines a Q-conditioned diffusion actor with a chunk-level critic for learning from mixed-quality deployment experience. The critic extends value estimation from individual actions to temporally extended action sequences.

2) We introduce future latent dynamics prediction as dense auxiliary supervision for chunk-level value learning, improving the critic's ability to distinguish beneficial and detrimental behaviors under sparse task rewards. The learned critic further provides Q-value estimates for constructing discrete quality conditions that guide diffusion-policy post-training.

3) Extensive simulation and real-world experiments demonstrate that PACL consistently improves pretrained policies and outperforms strong imitation-learning and offline-RL baselines. Ablation studies further confirm that both the Q-conditioned actor and predictive critic contribute positively to the overall improvement.

\section{Related work}


Imitation learning from mixed-quality data generally seeks to emphasize reliable segments while suppressing undesirable behaviors. Existing approaches mainly follow two strategies: weighting and filtering. Weighting-based methods estimate the reliability of transitions or demonstrators and assign greater influence to higher-quality data \cite{wu2019imitation,wang2021learning,xu2022discriminator}. Beyond sample-wise weighting, some methods model data quality more structurally through state-dependent expertise or distribution correction. Beliaev et al.\ estimate demonstrator expertise as a function of state \cite{beliaev2022expertise}, while Kim et al.\ use limited expert demonstrations to correct the occupancy distribution induced by a larger imperfect dataset \cite{kim2022demodice}. Filtering-based methods construct quality criteria to identify useful segments and exclude behaviors that are likely to degrade policy learning \cite{kuhar2023discern,wang2023pubc}. Yue et al.\ evaluate behavior quality through future outcomes, allowing useful behaviors to be retained even when their actions differ from expert demonstrations \cite{yue2024diverse}. In robotic manipulation, SSDF first learns self-supervised trajectory representations and scores failed segments by their similarity to expert behaviors, retaining high-quality segments for weighted imitation learning \cite{wu2025learning}. S2I instead segments demonstrations semantically, selects high-quality segments through contrastive representation learning, and refines suboptimal segments before policy training \cite{chen2025s2i}. Despite better utilizing mixed-quality experience, these methods primarily recover cleaner positive supervision. Failure-inducing segments are commonly discarded, downweighted, or replaced by corrected targets, leaving their negative evidence and delayed consequences largely unused during post-training.


Offline RL provides a complementary way to reuse mixed-quality experience by propagating outcomes from successful and failed rollouts through value learning. However, applying offline RL to visuomotor tasks remains difficult, and representative methods can substantially underperform imitation policies even with paired successful and failed trajectories \cite{mandlekar2022what}. On the actor side, diffusion-based RL improves policy expressiveness and policy extraction by modeling multimodal action distributions while preserving the behavior prior of offline data. DQL directly couples diffusion policy optimization with Q-maximization, whereas IDQL samples from a diffusion behavior policy and uses the critic to extract higher-value actions \cite{wang2022diffusion,hansen2023idql}. V-GPS similarly uses a value function learned through offline RL to rerank actions sampled from pretrained robot policies at deployment time, while keeping the proposal policy fixed \cite{nakamoto2025steering}. Complementary efforts focus on improving critic learning itself. Q-chunking extends value estimation to temporally extended actions \cite{li2025reinforcement}, while AC3 directly learns continuous action chunks under sparse rewards using intra-chunk returns and self-supervised intrinsic rewards to stabilize critic learning \cite{yang2026actor}. WCM further incorporates observation history and future latent prediction to strengthen temporal value representations \cite{fei2026wcm}. These developments address different bottlenecks in offline RL, but they do not directly resolve the challenge of learning from naturally collected deployment data. Most closely related to our setting, RISE uses Lipschitz regularization and distance-based augmentation to broaden local action support, enabling non-expert behaviors to be stitched back to the expert manifold \cite{huang2025rise}. However, this mechanism relies on sufficient local coverage between expert and non-expert data and on purposefully collected experience that supports such connections. For naturally accumulated deployment rollouts, sparse rewards and uncontrolled data coverage can still make value estimation and policy improvement unreliable.

\begin{figure*}[!t]
  \centering
  \includegraphics[width=0.90\textwidth]{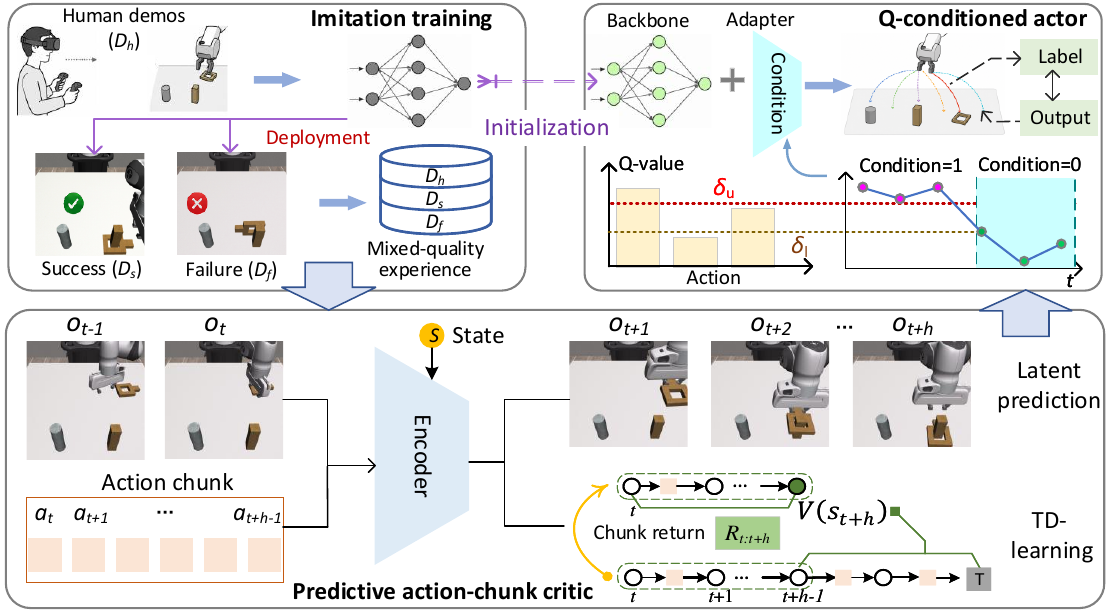}
  \caption{Overview of Predictive Action Chunk Learning (PACL). PACL improves a pretrained policy from human demonstrations and mixed-quality deployment rollouts using a Q-conditioned diffusion actor and a predictive chunk-level critic. The critic evaluates temporally extended actions through TD learning augmented with future latent prediction.}
  \label{fig:framework}
\end{figure*}

To obtain stronger learning signals beyond offline data, recent post-deployment methods introduce additional supervision or interaction during deployment. One line of work converts deployment failures into targeted supervision through human intervention. Hu et al. structure each intervention as a recovery segment that returns the robot to a familiar state, followed by a corrective segment for imitation-based finetuning \cite{hu2026rac}. Such data provides retry and adaptation behaviors largely absent from successful demonstrations. A second line uses outcome feedback to distinguish beneficial and detrimental behaviors. $\pi^{*}_{0.6}$ trains a distributional state-value function from sparse episode outcomes, uses multi-step value differences to assign binarized advantage labels to action chunks, and conditions a VLA policy on the resulting indicator \cite{physicalintelligence2025pi}. More interaction-intensive approaches optimize pretrained policies through newly collected on-robot rollouts and reward feedback, ranging from rollout-based policy optimization to iterative offline-to-online reinforcement learning \cite{tan2025interactive,lei2025rl100, wang2026learning}. While effective, these approaches depend on human supervision or physical interaction in the operational environment. In contrast, we focus on improving the policy from experience naturally accumulated during ordinary autonomous deployment, without deliberately collecting corrections or exploratory rollouts.

\section{Methods}

\subsection{Problem Formulation}
\label{sec:problem_formulation}

We consider episodic visual manipulation, where $o_t$ denotes the visual observation, $s_t$ denotes the robot proprioceptive state and $a_t$ denotes the robot action at time step $t$. We adopt Diffusion Policy (DP) \cite{chi2025diffusion} as the visuomotor policy $\pi_\theta$. Given the most recent $H_o$ observations, the policy generates an action chunk $A_t$ containing $H_a$ consecutive actions:
\begin{equation}
\nonumber
    A_t := (a_t,\ldots,a_{t+H_a-1}), \qquad
    A_t \sim \pi_\theta(\cdot \mid o_{t-H_o+1:t}, s_t),
\end{equation}
where $H_o$ and $H_a$ denote the observation and action horizons, respectively.
As illustrated in Fig.~\ref{fig:framework}, the initial policy $\pi_{\theta_0}$ is trained by imitation learning on a
human demonstration dataset $\mathcal{D}_h$. Each trajectory (i.e., rollout) is denoted by
$\tau_i=\{(o_t,s_t,a_t)\}_{t=0}^{T_i-1}$, where $T_i$ is the trajectory length.
After training, $\pi_{\theta_0}$ is deployed to autonomously collect
additional trajectories. A deployment trajectory is labeled successful if
the task is completed within the maximum episode horizon $T_{\max}$ and
failed otherwise. Let $y_i\in\{0,1\}$ denote this binary outcome. The
successful and failed deployment datasets are defined as
\begin{equation}
\nonumber
    \mathcal{D}_s = \{\tau_i \mid y_i=1\}, \quad
    \mathcal{D}_f = \{\tau_i \mid y_i=0\}.
\end{equation}
The complete dataset used for post-training is
$\mathcal{D}=\mathcal{D}_h\cup\mathcal{D}_s\cup\mathcal{D}_f$.

All these deployment trajectories are generated autonomously, without human intervention or corrective demonstrations. Their task outcomes provide the supervision used to construct the sparse rewards for critic learning. During the post-training, the diffusion actor is initialized from the pretrained policy $\pi_{\theta_0}$ rather than trained from scratch. Given the fixed dataset $\mathcal{D}$, PACL then performs critic learning and actor post-training entirely offline, as summarized in Fig.~\ref{fig:framework}.

\subsection{Predictive Action-Chunk Critic}
\label{sec:predictive_chunk_critic}

To exploit the mixed-quality experience in $\mathcal{D}$, we train a critic to estimate the long-term utility of generated action chunks.
Representative diffusion-based offline RL methods such as IDQL and RISE formulate value learning over transition-level state--action pairs \cite{hansen2023idql,huang2025rise}. This creates a temporal mismatch with diffusion actors, which naturally generate coordinated action sequences over an extended horizon. We therefore define the Q-function directly over the complete action chunk $A_t$, enabling long-horizon value estimation of temporally extended actions.

Given the task reward $r_t$, the discounted reward associated with an action
chunk is
\begin{equation}
    R_t =
    \sum_{k=0}^{K_t-1}\gamma^k r_{t+k},
    \qquad
    K_t=\min(H_a,T_i-t),
\end{equation}
where $\gamma$ is the discount factor and $K_t$ accounts for chunks truncated
near the end of a trajectory.
The observation history is encoded by the image backbone and fused with $s_t$ to obtain the latent state:
\begin{equation}
z_t=f_\xi(o_{t-H_o+1:t},s_t),
\end{equation}
where $f_\xi$ denotes the critic encoder. Based on this, we learn a value function $V_\psi(z_t)$
together with twin chunked Q-functions
$Q_{\phi_1}(z_t,A_t)$ and $Q_{\phi_2}(z_t,A_t)$. 
The value function is fitted
to an upper expectile of the target Q-values:
\begin{equation}
    \mathcal{L}_{V}
    =
    \mathbb{E}_{\mathcal{D}}
    \left[
        L_{\tau}
        \left(
            \min_{j=1,2}Q_{\bar{\phi}_j}(z_t,A_t)
            -
            V_{\psi}(z_t)
        \right)
    \right],
\end{equation}
where $L_{\tau}$ denotes the expectile regression loss and
$Q_{\bar{\phi}_j}$ are slowly updated target networks. 
The chunked temporal difference target and Q-function loss are
\begin{equation}
\begin{aligned}
    Y_t &=
    R_t+(1-d_t)\gamma^{K_t}V_\psi(z_{t+K_t}),\\
    \mathcal{L}_{Q}
    &=
    \sum_{j=1}^{2}
    \mathbb{E}_{\mathcal{D}}
    \left[
        \left(Q_{\phi_j}(z_t,A_t)-Y_t\right)^2
    \right],
\end{aligned}
\end{equation}
where $d_t$ indicates whether the episode terminates within the chunk. The encoder $f_\xi$ is jointly optimized with value functions.

Crucially, extending value estimation from individual actions to action chunks substantially increases the action dimensionality, while critic learning still relies mainly on scalar TD-target supervision. Under sparse rewards, this signal can be insufficient to distinguish action sequences with different future consequences. We therefore introduce an auxiliary latent dynamics objective that exploits the richer supervision contained in future observations.
For each action chunk, we randomly sample a prediction horizon
$h\in\{1,\ldots,K_t\}$. To preserve temporal causality, only the first $h$
actions are provided to the dynamics predictor $F_\eta$, while subsequent actions
are masked. Alongside the online encoder $f_\xi$, we maintain a target
encoder $f_{\bar{\xi}}$ updated by exponential moving average. The target
latent change and its prediction are defined as
\begin{equation}
\begin{aligned}
    \Delta\bar{z}_t^{(h)}
    &=
    f_{\bar{\xi}}
    (o_{t+h-H_o+1:t+h},s_{t+h})
    -
    f_{\bar{\xi}}
    (o_{t-H_o+1:t},s_t),\\
    \widehat{\Delta z}_t^{(h)}
    &=
    F_\eta(z_t,\widetilde{A}_t^{(h)},h),
\end{aligned}
\end{equation}
where $\widetilde{A}_t^{(h)}$ retains the first $h$ actions of $A_t$ and
masks the remaining actions. The latent dynamics loss is
\begin{equation}
    \mathcal{L}_{\mathrm{dyn}}
    =
    \mathbb{E}_{\mathcal{D},h}
    \left[
        \left\|
        \widehat{\Delta z}_t^{(h)}
        -
        \Delta\bar{z}_t^{(h)}
        \right\|^2
    \right].
\end{equation}
The target encoder provides a stable prediction target, while the online encoder is jointly optimized with the critic and dynamics model. This objective provides dense supervision across different prediction horizons and encourages action-sensitive representations for value estimation. The overall critic objective is
\begin{equation}
    \mathcal{L}_{\mathrm{critic}}
    =
    \mathcal{L}_{Q}
    +
    \lambda_V\mathcal{L}_{V}
    +
    \lambda_{\mathrm{dyn}}\mathcal{L}_{\mathrm{dyn}}.
\end{equation}

\begin{figure}[!htbp]
  \centering
    \begin{subfigure}[b]{0.495\linewidth}
      \includegraphics[width=\textwidth]{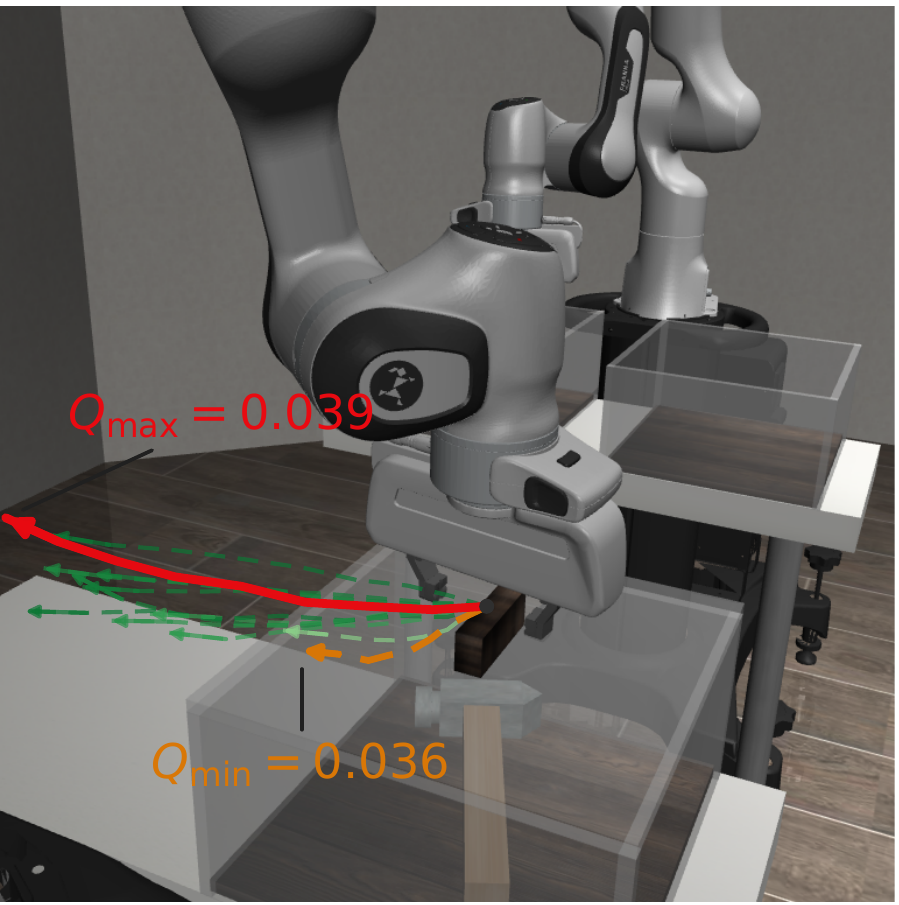}
      \caption{Transport}
    \end{subfigure}
    \begin{subfigure}[b]{0.49\linewidth}
      \includegraphics[width=\textwidth]{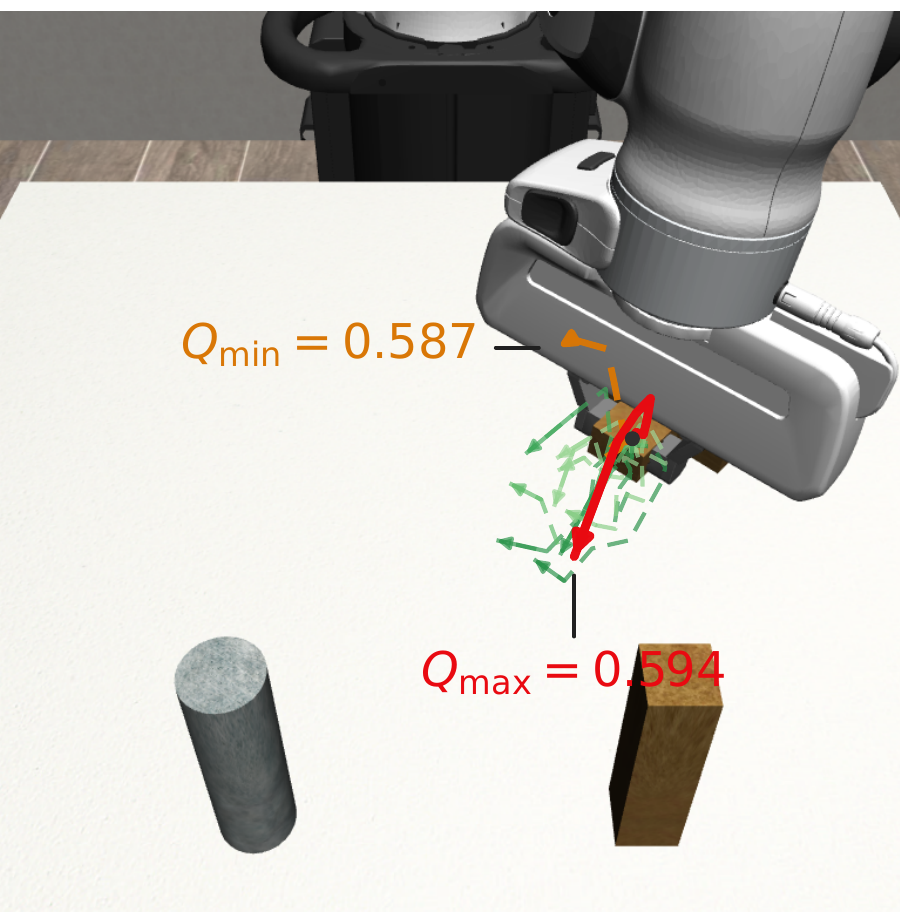}
      \caption{Square}
    \end{subfigure}
  \caption{Visualization of critic-based action chunk ranking. The predictive critic evaluates multiple candidate action chunks and selects the highest-valued one for execution on Transport and Square tasks.}
  \label{fig:q_selection}
\end{figure}

As illustrated in Fig.~\ref{fig:q_selection}, candidate action chunks generated from the same observation can lead to different future behaviors. The predictive critic assigns higher Q-values to chunks that better advance task completion, enabling the actor to select the most promising candidate for execution.

\subsection{Q-Conditioned Actor Learning}
\label{sec:condition_actor}

\begin{figure}[!htbp]
  \centering
      \begin{subfigure}[b]{\linewidth}
      \includegraphics[width=0.492\linewidth]{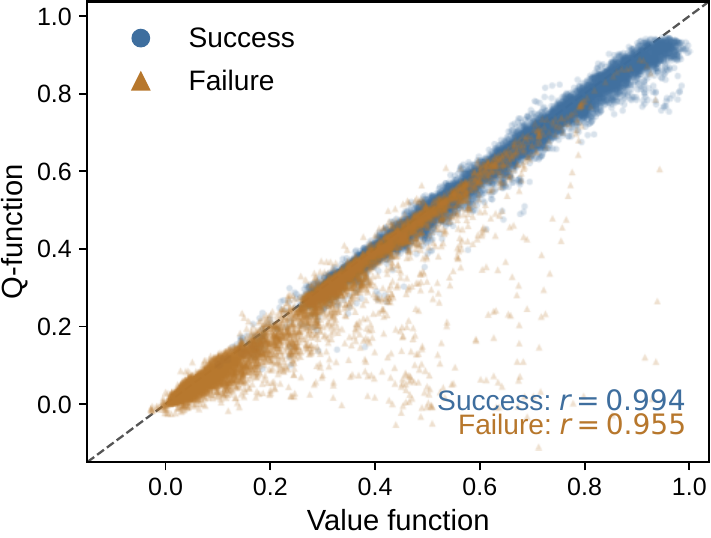}
      \includegraphics[width=0.492\linewidth]{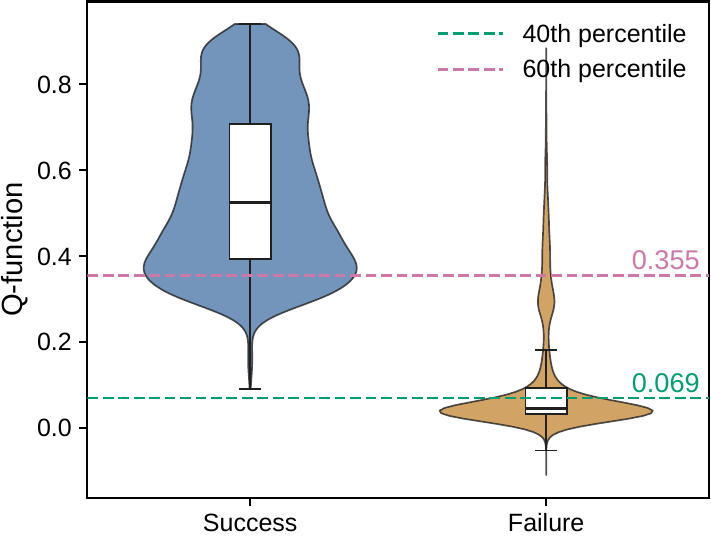}
      \end{subfigure}
  \caption{Analysis of critic estimates on Square. Left: state values and chunk-level Q-values are strongly correlated for both successful and failed rollouts. Right: Q-value distributions show clear separation between successful and failed rollouts.}
  \label{fig:q_function_dist}
\end{figure}

Directly incorporating low-quality deployment actions into an imitation objective may reinforce undesirable behaviors. We therefore use the learned critic to construct quality conditions for actor post-training. As shown in Fig.~\ref{fig:q_function_dist}, the state value $V$ and
chunk-level $Q$ exhibit strongly correlated trends across states, with
Pearson correlations of $0.994$ and $0.955$ on successful and failed
Square rollouts, respectively. This strong correlation is consistent
with IQL-style learning, where $V$ is fitted to an expectile of $Q$.
Meanwhile, the $Q$-value distributions show clear separation between
successful and failed rollouts. We therefore directly use chunk-level
$Q$ values to construct discrete quality conditions.

For each action chunk $A_t$ in $\mathcal{D}_s \cup \mathcal{D}_f$, we evaluate its conservative chunk value as
\begin{equation}
q_t = \min_{j=1,2} Q_{\phi_j}(z_t,A_t).
\end{equation}
Let $\delta_l$ and $\delta_u$ denote the lower and upper percentile thresholds of the $Q$-value distribution over deployment experience. The quality condition is assigned as
\begin{equation}
c_t =
\begin{cases}
1, & q_t \geq \delta_u, \\
0, & q_t \leq \delta_l, \\
\varnothing, & \text{otherwise},
\end{cases}
\end{equation}
where $\varnothing$ denotes an uncertain quality condition. We use the 40th and 60th percentiles for $\delta_l$ and $\delta_u$, respectively, leaving an abstention region for chunks with ambiguous critic estimates.

During post-training, we introduce a lightweight condition adapter to the pretrained DP, which is the only architectural modification to the actor and maps $c_t$ to a conditioning embedding. Samples with $c_t=\varnothing$ use a null condition and are trained unconditionally, while all human demonstrations in $\mathcal{D}_h$ are assigned $c_t=1$ to preserve reliable demonstrated behaviors. We then finetune the pretrained DP on $\mathcal{D}_h \cup \mathcal{D}_s \cup \mathcal{D}_f$ using its original denoising objective conditioned on $c_t$:
\begin{equation}
\nonumber
\mathcal{L}_{\mathrm{actor}}
=
\mathbb{E}_{(o_t,A_t,s_t)\sim\mathcal{D}}
\left[
\left\|
\epsilon -
\epsilon_\theta
\left(
A_t^k,k,o_{t-H_o+1:t},s_t,c_t
\right)
\right\|^2
\right],
\end{equation}
where $A_t^k$ denotes the noisy action chunk at diffusion step $k$, and $\epsilon$ is the injected noise.

Rather than directly optimizing the actor with critic values, PACL distills critic estimates into discrete Q-conditions for diffusion-policy post-training. In IQL-style offline critic learning, $V$ is obtained through expectile regression over $Q$, and the two estimates can therefore exhibit strongly correlated trends across dataset states, as observed in Fig.~\ref{fig:q_function_dist}. We thus use the chunk-level $Q$ directly to construct simple percentile-based conditions, instead of forming conditions from $Q-V$ as in advantage-conditioned methods \cite{fei2026wcm}. This design is particularly suitable for naturally accumulated deployment experience, as it requires only sparse task outcomes and does not rely on additional action-quality annotations. It also differs from IDQL \cite{hansen2023idql} and V-GPS \cite{nakamoto2025steering}, where the critic does not shape the proposal distribution during actor post-training and is primarily used for action selection at inference time. In PACL, the critic guides both offline actor post-training through Q-conditioning and candidate selection during deployment, enabling the complete mixed-quality dataset to be exploited without explicit behavior filtering.

At inference, the condition is fixed to $c_t=1$, and the actor samples $N$ candidate action chunks from the high-quality conditional distribution. The predictive critic evaluates the candidates and selects the highest-valued chunk for execution. After executing it for $H_a$ steps, the policy receives a new observation and repeats the process.

\section{Experiments}

\subsection{Setup}

\begin{figure}[!ht]
\captionsetup[subfigure]{labelformat=empty}
  \centering
    \begin{subfigure}[b]{0.245\linewidth}
      \includegraphics[width=\textwidth]{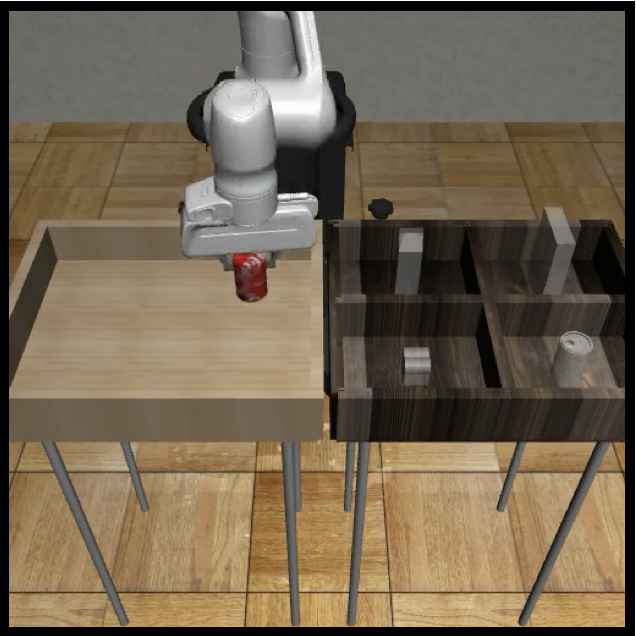}
      \caption{Can}
    \end{subfigure}%
    \hfill
    \begin{subfigure}[b]{0.245\linewidth}
      \includegraphics[width=\textwidth]{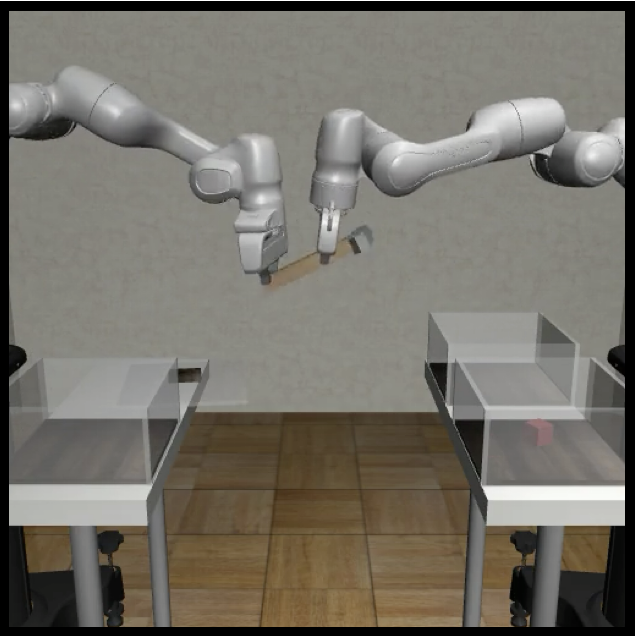}
      \caption{Transport}
    \end{subfigure}%
    \hfill
    \begin{subfigure}[b]{0.245\linewidth}
      \includegraphics[width=\textwidth]{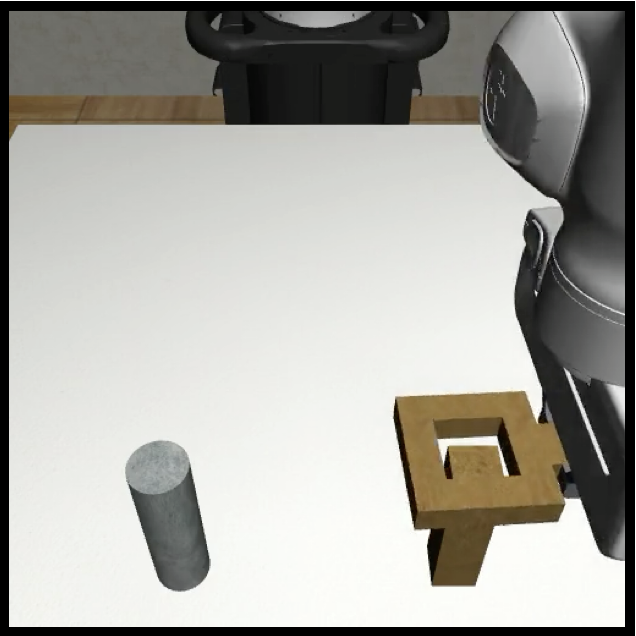}
      \caption{Square}
    \end{subfigure}%
    \hfill
    \begin{subfigure}[b]{0.245\linewidth}
      \includegraphics[width=\textwidth]{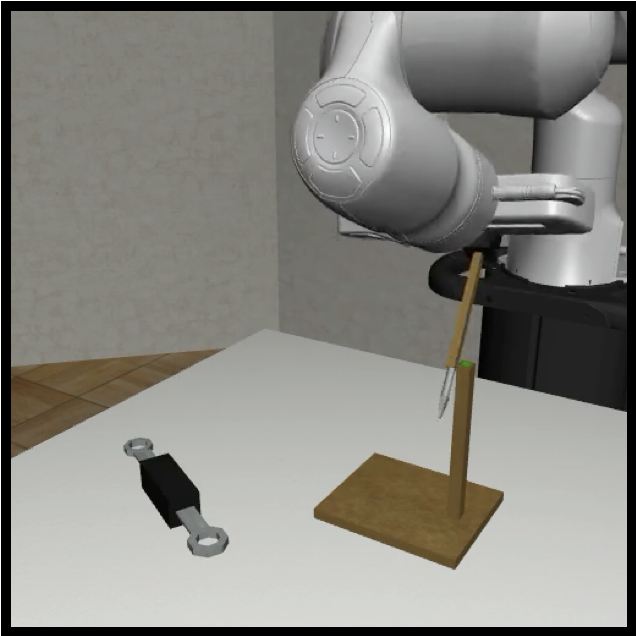}
      \caption{ToolHang}
    \end{subfigure}

  \begin{subfigure}[b]{0.32\linewidth}
    \includegraphics[width=1.0\textwidth,trim={0cm 0cm 0cm 0cm},clip]{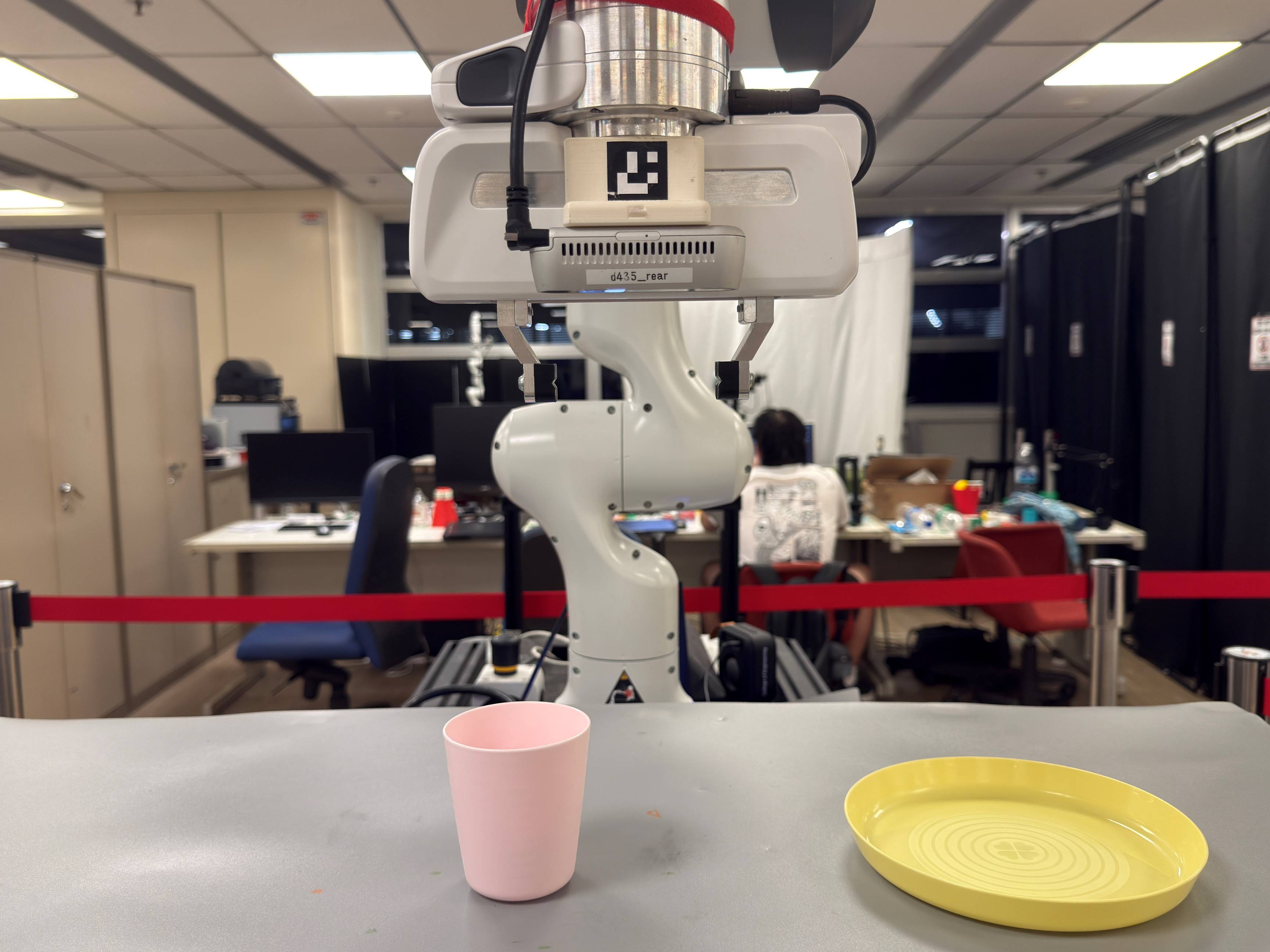}
    \caption{PickCup}
  \end{subfigure}
  \begin{subfigure}[b]{0.32\linewidth}
    \includegraphics[width=1.0\textwidth,trim={0cm 0cm 0cm 0cm},clip]{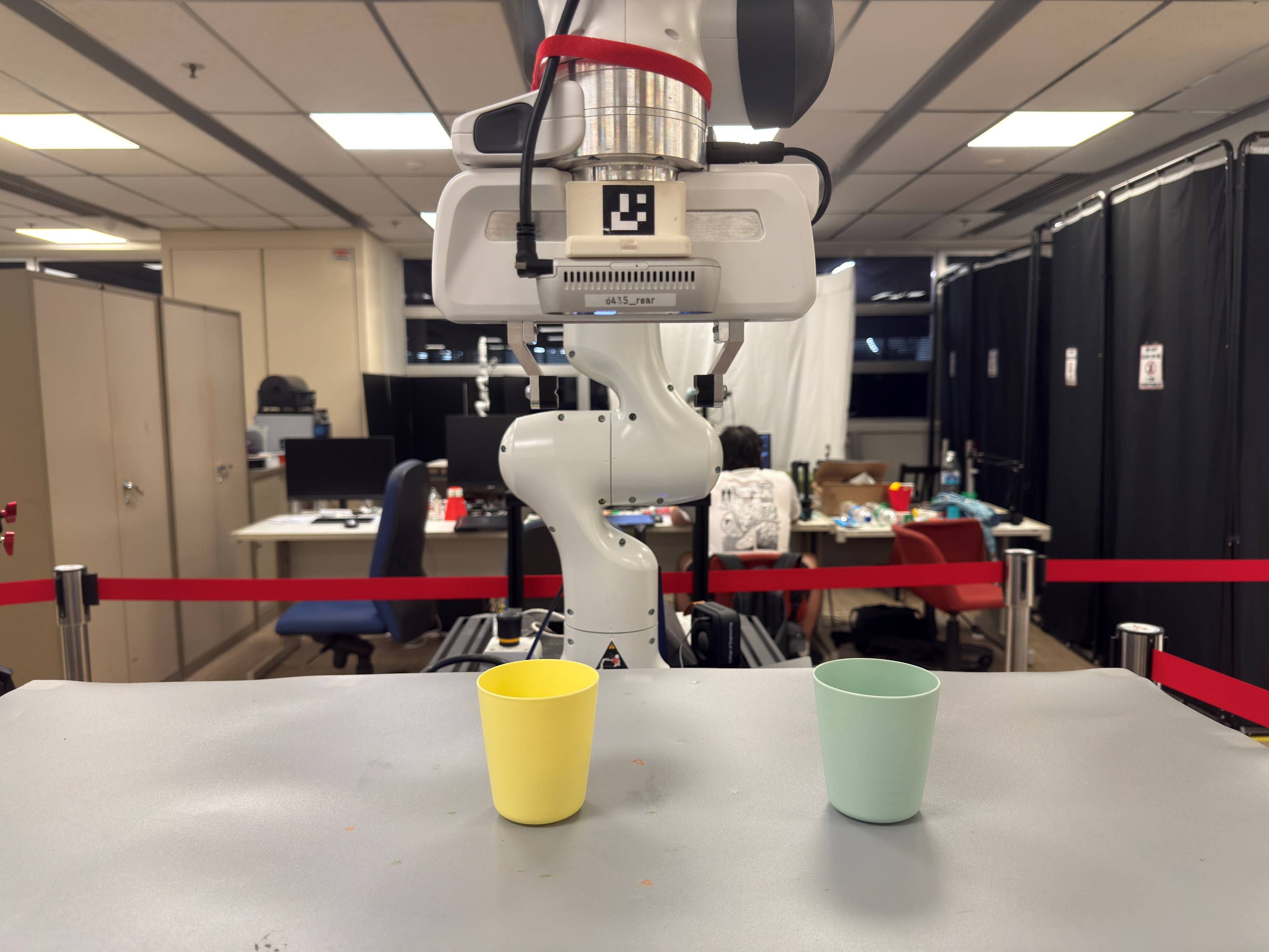}
    \caption{StackCup}
  \end{subfigure}
  \begin{subfigure}[b]{0.32\linewidth}
    \includegraphics[width=1.0\textwidth,trim={0cm 0cm 0cm 0cm},clip]{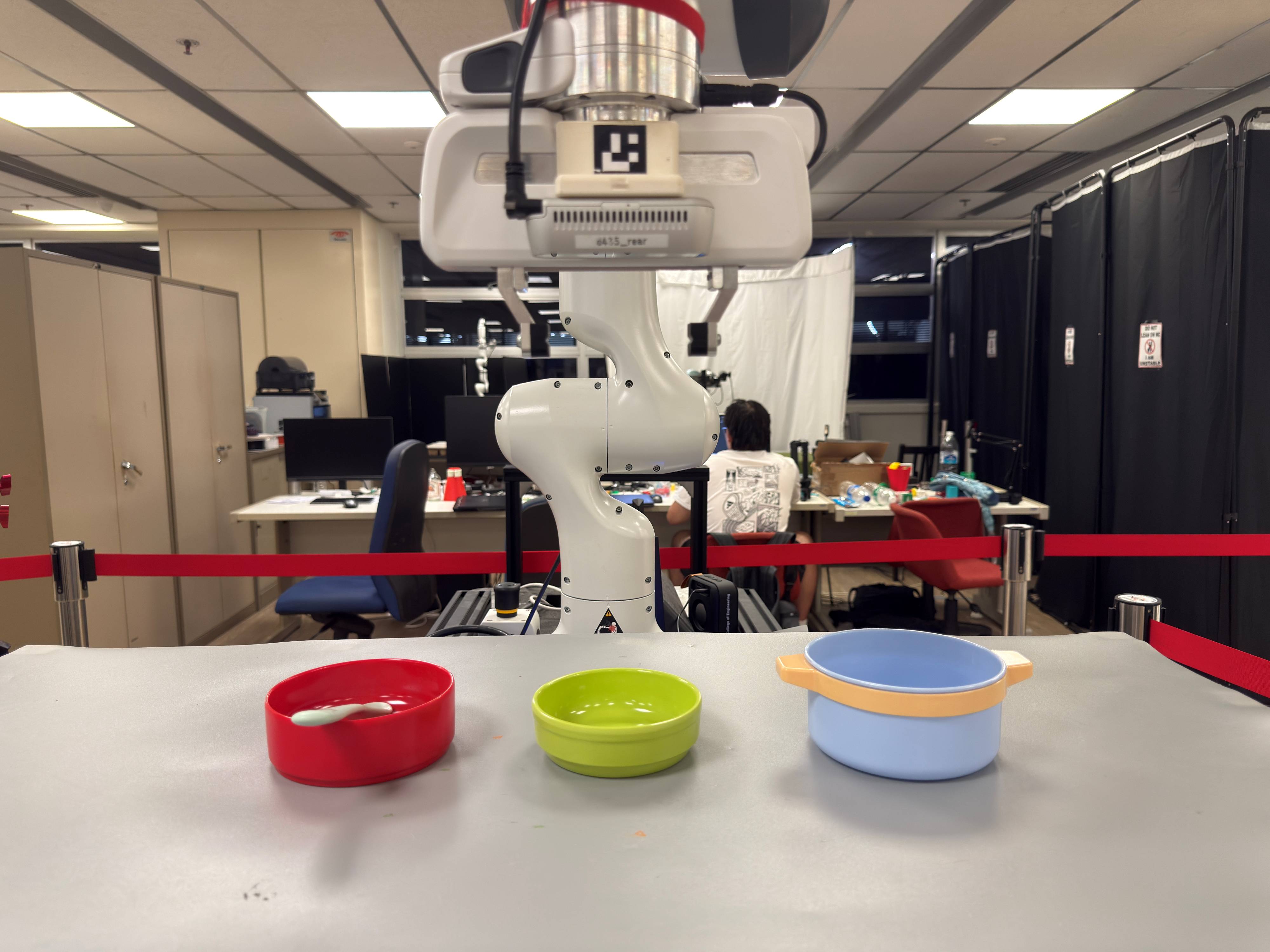}
    \caption{MoveSpoon}
  \end{subfigure}
  \caption{Experiment setup. This includes 4 simulation tasks on the \textit{Robomimic} benchmark and 3 real-robot tasks on the Franka arm.}
  \label{fig:experiment_task}
\end{figure}

\textbf{Simulation experiments.} Our simulation experiments are based on the Robomimic benchmark \cite{mandlekar2022what}, which provides standardized robotic manipulation tasks together with human demonstration datasets. As shown in Fig.~\ref{fig:experiment_task}, we evaluate our method on four visual manipulation tasks with increasing complexity: \textit{Can}, \textit{Transport}, \textit{Square} and \textit{ToolHang}. \textit{Can} requires the robot to grasp a can and place it into the target bin.  \textit{Transport} is a long-horizon task involving multi-stage object transport, and \textit{Square} requires grabbing and inserting a square nut onto the corresponding peg. \textit{ToolHang} requires assembling the hanging structure and placing the tool onto it. Task horizons $T_{\max}$ are 400 steps for \textit{Can} and \textit{Square}, and 700 steps for \textit{Transport} and \textit{ToolHang}. For each task, we use 200 human demonstrations to train a visuomotor Diffusion Policy as the initial baseline for deployment. \textit{Can} is trained for 50 epochs and other tasks are trained for 200 epochs. The resulting policies are then deployed autonomously to collect 500 rollouts of mixed experience for post-training.

\begin{table*}[!th]
    \centering
    \caption{
    Post-deployment improvement across 4 simulation and 3 real-world manipulation tasks. All methods are warm-started from the pretrained DP and evaluated over 250 simulation rollouts or 25 real-world trials per task.
    }
    \label{tab:main_results}
    \begin{tabular}{cll|cccc|ccc}
        \toprule
        \multicolumn{2}{c}{Method}
        & Data source
        & Can
        & Transport
        & Square
        & ToolHang
        & PickCup
        & StackCup
        & MoveSpoon \\
        \midrule

        Baseline
        & DP~\cite{chi2025diffusion}
        & $D_{h}$
        & 88.0
        & 84.0
        & 78.8
        & 46.4
        & 84.0
        & 72.0
        & 64.0 \\
        \midrule

        \multirow{3}{*}{IL}
        & SUB \cite{mirchandani2025scale}
        & $D_{h} + D_{s} + D_{f}$
        & {93.6}
        & {88.0}
        & {80.8}
        & {70.0}
        & {92.0}
        & {84.0}
        & {68.0} \\

        & Self-Imitation
        & $D_{h} + D_{s}$
        & {94.8}
        & {89.2}
        & {86.8}
        & {78.0}
        & \textbf{100.0}
        & {92.0}
        & {72.0} \\

        & SSDF~\cite{wu2025learning}
        & $D_{h} + D_{s} + D_{f}^{+}$
        & \textbf{98.4}
        & {92.0}
        & {88.4}
        & {81.2}
        & \textbf{100.0}
        & {96.0}
        & {80.0}\\
        \midrule

        \multirow{4}{*}{\makecell{Offline\\RL}}
        & DQL~\cite{wang2022diffusion}
        & $D_{h} + D_{s} + D_{f}$
        & {92.0}
        & {86.4}
        & {81.6}
        & {56.4}
        & {88.0}
        & {76.0}
        & {72.0}
        \\

        & IDQL~\cite{hansen2023idql}
        & $D_{h} + D_{s} + D_{f}$
        & {91.6}
        & {90.0}
        & {83.6}
        & {58.8}
        & \textbf{100.0}
        & {92.0}
        & {76.0}
        \\

        & RISE~\cite{huang2025rise}
        & $D_{h} + D_{s} + D_{f}$
        & {94.0}
        & {90.8}
        & {86.0}
        & {64.4}
        & \textbf{100.0}
        & {96.0}
        & {80.0}
        \\
        \cmidrule(l){2-10}
        & PACL
        & $D_{h} + D_{s} + D_{f}$
        & \textbf{98.4}
        & \textbf{96.0}
        & \textbf{93.2}
        & \textbf{82.8}
        & \textbf{100.0}
        & \textbf{100.0}
        & \textbf{84.0} \\

        \bottomrule
    \end{tabular}
\end{table*}

\textbf{Real-robot experiments.} We use a Franka Panda 7-DoF robot arm with two Intel RealSense cameras providing third-person and wrist-mounted RGB observations. Images from both views are resized to $224\times224$, encoded by the visual backbone, and fused with the robot proprioceptive state as policy input. Human demonstrations are collected via teleoperation, and the policy outputs 7-DoF actions consisting of a 6-DoF end-effector displacement and a gripper command at 20 Hz. We evaluate three manipulation tasks: \textit{PickCup}, \textit{StackCup}, and \textit{MoveSpoon}. The same DP architecture is first trained from human demonstrations and then deployed autonomously to collect additional rollouts for post-training. During closed-loop execution, the policy asynchronously predicts 8-step action chunks for robot control. For each task, we collect 50 human demonstrations for initial baseline training and 50 autonomous rollouts for post-training.

\textbf{Baselines and data sources.}
We compare PACL with three imitation-learning and three offline RL baselines. For imitation learning, suboptimal behavior cloning (SUB) directly finetunes the pretrained policy on all collected trajectories without quality distinction \cite{mirchandani2025scale}. \textit{Self-Imitation} uses only successful deployment rollouts together with human demonstrations, and SSDF \cite{wu2025learning} selects useful segments from failed rollouts for weighted imitation learning. For offline RL, we include DQL \cite{wang2022diffusion}, IDQL \cite{hansen2023idql}, and RISE \cite{huang2025rise}, all trained on the full mixed-quality dataset. We denote the original human demonstrations by $\mathcal{D}_h$, successful rollouts by $\mathcal{D}_s$, and failed rollouts by $\mathcal{D}_f$. $\mathcal{D}_f^{+}$ denotes the high-quality segments extracted from failed rollouts by SSDF. All these post-training methods are warm-started from the same pretrained Diffusion Policy.

\textbf{Implementation details.}
All models are trained on 8 NVIDIA A100 GPUs with a batch size of 100 per GPU. During post-training, the image backbone uses a learning rate of $1\times10^{-5}$, while the condition adapter and diffusion U-Net use $1\times10^{-4}$; both are decayed to zero with a cosine schedule. All post-training variants are trained for 50 epochs. The observation and action horizons are $H_o=2$ and $H_a=8$, respectively. For critic learning, the IQL expectile is set to $\tau=0.9$, with $\lambda_V=1.0$ and $\lambda_{\mathrm{dyn}}=0.5$. We use sparse terminal rewards: all intermediate rewards are zero, while the terminal reward is $1$ for human demonstrations and successful deployment rollouts and $0$ for failed rollouts. The discount factor is set to $\gamma=0.99$. At inference, DDIM with 100 denoising steps is used, and the critic selects the best action chunk from $N=12$ candidates.

\subsection{Results}

Table~\ref{tab:main_results} shows the comparison results of PACL with imitation-learning and offline RL baselines across simulation and real-robot tasks. Directly finetuning on all mixed-quality rollouts (SUB) provides only modest gains over the pretrained policy, while Self-Imitation is generally more effective by restricting supervision to successful rollouts. SSDF is the strongest imitation-learning baseline, benefiting from explicitly identifying and retaining useful segments from failed experience. Among offline-RL methods, DQL yields limited improvements and often remains below the stronger imitation baselines, whereas IDQL benefits more consistently from critic-guided policy extraction. RISE further improves performance by broadening local action support, but its gains remain constrained when naturally collected rollouts provide insufficient state-action coverage for effective stitching. In contrast, PACL achieves the strongest overall results, matching or surpassing SSDF across all reported tasks; for example, it reaches $96.0\%$ on Transport and $93.2\%$ on Square. Importantly, unlike filtering-based methods that require explicit extraction of useful segments from failed trajectories, PACL directly learns from the complete mixed-quality dataset using only simple trajectory-level outcome rewards, providing a more automated route for post-deployment policy improvement.

\subsection{Ablation study}

\begin{figure}[!thpb]
  \centering
  \includegraphics[width=0.48\textwidth]{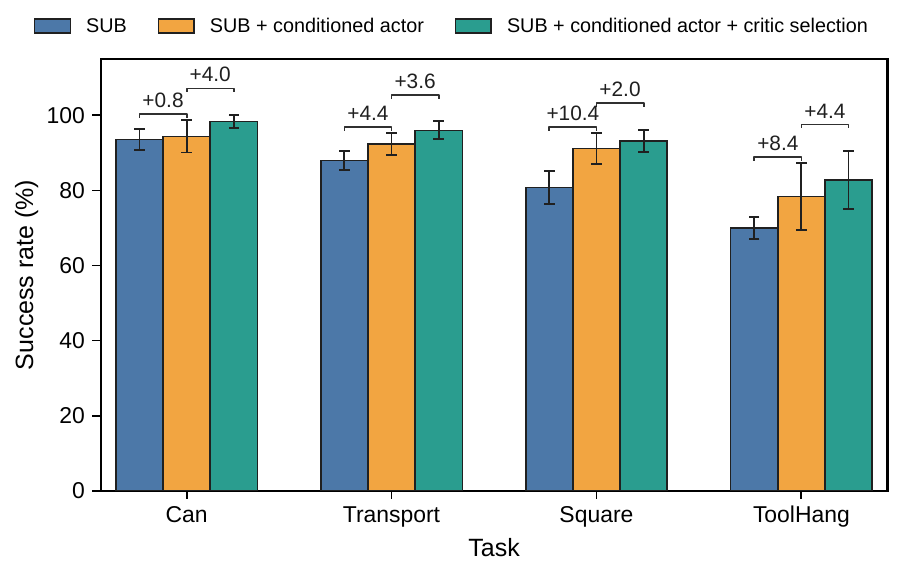}
  \caption{Component ablation of PACL. Both the Q-conditioned actor and predictive action-chunk critic consistently improve performance across all simulation tasks.}
  \label{fig:ablation_action_critc}
\end{figure}

\paragraph{Impact of the Q-conditioned actor and predictive action-chunk critic}
We progressively add the two key components of PACL to the SUB baseline. As shown in Fig.~\ref{fig:ablation_action_critc}, the Q-conditioned actor consistently improves performance across all four tasks, with particularly clear gains on Square and ToolHang. This shows that separating action chunks according to critic-estimated quality enables more effective learning from mixed-quality deployment experience than unconditional supervision. Using the predictive critic additionally for candidate selection provides further improvements on every task. These results demonstrate complementary benefits from the two components: the quality conditioning improves actor post-training, while the predictive critic further refines action selection by ranking candidate chunks before execution.

\paragraph{Ablation of critic design}
We further isolate the effect of critic design on Transport by freezing the pretrained DP and using each critic to rank the same $N$ sampled action chunks. As shown in Table~\ref{tab:ablation_critic}, the standard one-step critic degrades rapidly as $N$ increases, indicating unreliable value ranking over larger candidate sets. Adding next-frame latent prediction substantially improves the one-step critic, demonstrating the benefit of predictive visual supervision for value learning. Chunk-level evaluation alone provides limited and inconsistent improvement, whereas combining it with future latent prediction in PACL achieves the best performance for these four candidate-set sizes. These results show that predictive supervision substantially improves critic reliability, while action-chunk evaluation provides complementary benefits for ranking temporally extended behaviors.

\begin{table}[!htbp]
    \centering
    \caption{
    Ablation of critic design on Transport.
    }
    \label{tab:ablation_critic}
    \begin{tabular}{lcccc}
        \toprule
        \textbf{Method} 
        & \textbf{N=4} 
        & \textbf{N=8} 
        & \textbf{N=12} 
        & \textbf{N=16} \\
        \midrule

        One-step critic
        & 79.2 
        & 75.2
        & 63.6
        & 58.8 \\

        Predictive one-step critic
        & 85.2
        & 82.8
        & 83.2
        & 76.4 \\

        Chunk critic
        & 85.2
        & 73.2
        & 65.2
        & 63.6 \\

        PACL
        & 89.2
        & 83.2
        & 86.8
        & 80.4 \\
        
        
        \bottomrule
    \end{tabular}
\end{table}

\begin{figure}[!htbp]
  \centering
  \includegraphics[width=0.85\linewidth]{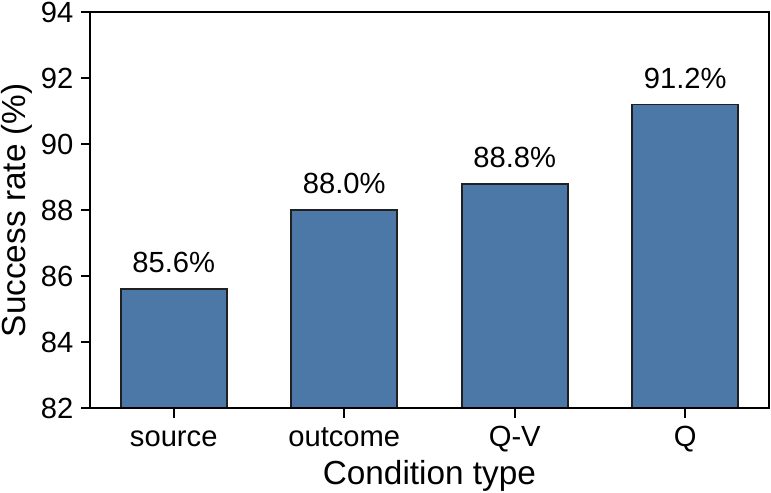}
  \caption{Comparison of conditioning strategies for actor post-training with single-sample inference ($N=1$).}
\label{fig:actor_ablation}
\end{figure}

\paragraph{Ablation of conditioning strategy}
In Fig.~\ref{fig:actor_ablation}, we compare four conditioning strategies on Square with $N=1$, so that no critic-based candidate selection is involved at inference. \textit{Source Condition} assigns the positive condition only to human demonstrations and negative condition to all deployment data. \textit{Outcome Condition} assigns the positive condition to human and successful data. \textit{$Q-V$ Condition} constructs chunk-level conditions from the estimated advantage, whereas \textit{Q-Condition} uses the chunk-level Q-value directly. The corresponding success rates are $85.6\%$, $88.0\%$, $88.8\%$, and $91.2\%$, respectively. The results show that critic-derived chunk-level supervision is more effective than source- or outcome-level conditioning. Among the critic-based variants, direct Q-conditioning achieves the best performance, supporting the use of chunk-level Q estimates for actor post-training.

\paragraph{Impact of deployment data composition}
We study the effect of post-training data composition in Table~\ref{tab:ablation_data}. All variants load the same pretrained model and are trained for another 50 epochs. \textit{Post-trained DP} continues imitation learning using only the 200 human demonstrations; \textit{PACL-success} uses the demonstrations with 100 successful rollouts; \textit{PACL-small} further adds 50 failed rollouts; \textit{PACL-rollout} uses all 500 autonomous rollouts without human demonstrations; and \textit{PACL-full} uses the complete dataset. Continuing imitation learning yields only moderate gains. More importantly, when PACL is trained with successful rollouts only, the critic lacks negative experience for discriminative value learning, resulting in degraded overall performance. Introducing 50 failed rollouts improves Square from $77.6\%$ to $88.0\%$ and Transport from $83.2\%$ to $88.8\%$, showing that failure experience provides essential supervision for critic learning. PACL-small also outperforms the rollout-only variant despite using fewer trajectories, suggesting that human demonstrations remain a valuable anchor during post-training. Overall, the results highlight the complementary roles of expert demonstrations and mixed-quality deployment experience.

\begin{table}[!htbp]
    \centering
    \caption{
    Ablation of post-training data composition and scale. ``Data'' denotes the total number of trajectories used for training. $N=12$ candidate chunks are used at inference for all PACL variants.
    }
    \label{tab:ablation_data}
    \begin{tabular}{lc|ccc}
        \toprule
        \textbf{Setting} 
        & \textbf{Data/Epochs} 
        & \textbf{Square} 
        & \textbf{Transport} \\
        \midrule

        Baseline DP
        & 200 / 0 
        & 78.8
        & 84.0 \\

        Post-trained DP
        & 200 / 50
        & 84.0
        & 86.4 \\

        PACL-success
        & 300 / 50
        & 77.6
        & 83.2 \\

        PACL-rollout
        & 500 / 50
        & 87.6
        & 86.8 \\

        PACL-small
        & 350 / 50
        & 88.0
        & 88.8 \\
        
        PACL-full
        & 700 / 50
        & 93.2
        & 96.0 \\
        
        \bottomrule
    \end{tabular}
\end{table}

\section{Conclusions}

We presented Predictive Action Chunk Learning (PACL) for improving robot manipulation policies from naturally accumulated mixed-quality deployment experience. PACL first learns a predictive action-chunk critic to evaluate temporally extended behaviors, and then distills its Q-values into discrete conditions for diffusion-policy post-training. Experiments show consistent improvements over strong imitation-learning and offline RL baselines, while ablations confirm the importance of predictive critic learning, action-chunk evaluation, and failure experience. These results suggest that naturally collected deployment data can provide a practical basis for automated post-deployment robot learning.







\bibliographystyle{IEEEtran}
\bibliography{reference}

@article{chi2025diffusion,
  title={Diffusion policy: Visuomotor policy learning via action diffusion},
  author={Chi, Cheng and Xu, Zhenjia and Feng, Siyuan and others},
  journal={International Journal of Robotics Research},
  volume={44},
  number={10-11},
  pages={1684--1704},
  year={2025},
  publisher={Sage Publications Sage UK: London, England}
}

@inproceedings{wu2019imitation,
  title={Imitation learning from imperfect demonstration},
  author={Wu, Yueh-Hua and Charoenphakdee, Nontawat and Bao, Han and Tangkaratt, Voot and Sugiyama, Masashi},
  booktitle={International Conference on Machine Learning},
  year={2019},
  organization={PMLR}
}

@article{lesort2020continual,
  title={Continual learning for robotics: Definition, framework, learning strategies, opportunities and challenges},
  author={Lesort, Timoth{\'e}e and Lomonaco, Vincenzo and Stoian, Andrei and others},
  journal={Information fusion},
  volume={58},
  pages={52--68},
  year={2020},
  publisher={Elsevier}
}

@inproceedings{spencer2020learning,
  title     = {Learning from Interventions: Human-Robot Interaction
               as Both Explicit and Implicit Feedback},
  author    = {Spencer, Jonathan and Choudhury, Sanjiban and
               Barnes, Matt and others},
  booktitle = {Robotics: Science and Systems},
  year      = {2020},
  doi       = {10.15607/RSS.2020.XVI.055}
}

@inproceedings{wang2021learning,
  title={Learning to weight imperfect demonstrations},
  author={Wang, Yunke and Xu, Chang and Du, Bo and others},
  booktitle={International Conference on Machine Learning},
  year={2021},
  organization={PMLR}
}

@inproceedings{xu2022discriminator,
  title={Discriminator-weighted offline imitation learning from suboptimal demonstrations},
  author={Xu, Haoran and Zhan, Xianyuan and Yin, Honglei and others},
  booktitle={International Conference on Machine Learning},
  year={2022},
  organization={PMLR}
}

@inproceedings{beliaev2022expertise,
  title={Imitation learning by estimating expertise of demonstrators},
  author={Beliaev, Mark and Shih, Andy and Ermon, Stefano and others},
  booktitle={International Conference on Machine Learning},
  year={2022},
  organization={PMLR}
}

@inproceedings{kim2022demodice,
  title     = {{DemoDICE}: Offline Imitation Learning with Supplementary
               Imperfect Demonstrations},
  author={Kim, Geon-Hyeong and Seo, Seokin and Lee, Jongmin and others},
  booktitle={International Conference on Learning Representations},
  year={2022}
}

@inproceedings{kuhar2023discern,
  title={Learning to discern: Imitating heterogeneous human demonstrations with preference and representation learning},
  author={Kuhar, Sachit and Cheng, Shuo and Chopra, Shivang and others},
  booktitle={Conference on Robot Learning},
  year={2023},
  organization={PMLR}
}

@inproceedings{wang2023pubc,
  title={Improving behavioural cloning with positive unlabeled learning},
  author={Wang, Qiang and McCarthy, Robert and Bulens, David Cordova and others},
  booktitle={Conference on robot learning},
  year={2023},
  organization={PMLR}
}

@inproceedings{yue2024diverse,
  title     = {How to Leverage Diverse Demonstrations in Offline
               Imitation Learning},
  author    = {Yue, Sheng and Liu, Jiani and Hua, Xingyuan                 and others},
  booktitle = {International Conference on
               Machine Learning},
  year      = {2024},
  publisher = {PMLR}
}

@inproceedings{wu2025learning,
  title={Learning from imperfect demonstrations with self-supervision for robotic manipulation},
  author={Wu, Kun and Liu, Ning and Zhao, Zhen and others},
  booktitle={International Conference on Robotics and Automation},
  year={2025},
  organization={IEEE}
}

@inproceedings{chen2025s2i,
  title     = {Towards Effective Utilization of Mixed-Quality Demonstrations in Robotic Manipulation via Segment-Level Selection and Optimization},
  author    = {Chen, Jingjing and Fang, Hongjie and
               Fang, Hao-Shu and others},
  booktitle = {International Conference on Robotics and
               Automation},
  year      = {2025},
  organization = {IEEE}
}

@article{hansen2023idql,
  title={{IDQL}: Implicit {Q}-learning as an actor-critic method with diffusion policies},
  author={Hansen-Estruch, Philippe and Kostrikov, Ilya and Janner, Michael and others},
  journal={arXiv preprint arXiv:2304.10573},
  year={2023}
}

@inproceedings{wang2022diffusion,
  title={Diffusion policies as an expressive policy class for offline reinforcement learning},
  author={Wang, Zhendong and Hunt, Jonathan J and Zhou, Mingyuan},
  booktitle={International Conference on Learning Representations},
  year={2022}
}

@inproceedings{kostrikov2022offline,
  title     = {Offline Reinforcement Learning with Implicit {Q}-Learning},
  author    = {Kostrikov, Ilya and Nair, Ashvin and Levine, Sergey},
  booktitle = {International Conference on Learning Representations},
  year      = {2022}
}

@inproceedings{mandlekar2022what,
  title     = {What Matters in Learning from Offline Human                       Demonstrations for Robot Manipulation},
  author    = {Mandlekar, Ajay and Xu, Danfei and Wong, Josiah and                     others},
  booktitle = {Conference on Robot Learning},
  publisher = {PMLR},
  year      = {2022}
}

@article{huang2025rise,
  title         = {Using Non-Expert Data to Robustify Imitation Learning
                   via Offline Reinforcement Learning},
  author        = {Huang, Kevin and Scalise, Rosario and Winston, Cleah and others},
  journal       = {arXiv preprint arXiv:2510.19495},
  year          = {2025},
  eprint        = {2510.19495},
  archivePrefix = {arXiv},
}

@inproceedings{li2025reinforcement,
  title     = {Reinforcement Learning with Action Chunking},
  author    = {Li, Qiyang and Zhou, Zhiyuan and Levine, Sergey},
  booktitle = {Advances in Neural Information Processing Systems},
  year      = {2025}
}

@article{hu2026rac,
  title={{RAC}: Robot learning for long-horizon tasks by scaling recovery and correction},
  author={Hu, Zheyuan and Wu, Robyn and Enock, Naveen and others},
  journal={IEEE Transactions on Robotics},
  year={2026},
  publisher={IEEE}
}

@article{fei2026wcm,
  title={{WCM}: A World Critic Model for Vision-Language-Action Reinforcement Learning},
  author={Fei, Senyu and Yu, Xiaopeng and Wang, Siyin and Zhao, Xianzhong and Gong, Jingjing and Qiu, Xipeng},
  journal={arXiv preprint arXiv:2607.29613},
  year={2026}
}

@inproceedings{physicalintelligence2025pi,
  title     = {$\pi_{0.6}^{*}$: A {VLA} That Learns From Experience},
  author    = {Amin, Ali and Aniceto, Raichelle and Balakrishna, Ashwin and others},
  booktitle = {Robotics: Science and Systems},
  year      = {2026}
}

@inproceedings{
tan2025interactive,
title={Interactive Post-Training for Vision-Language-Action Models},
author={Shuhan Tan and Kairan Dou and Yue Zhao and Philipp Kraehenbuehl},
booktitle={Workshop on Foundation Models Meet Embodied Agents at CVPR},
year={2025}
}

@inproceedings{brohan2023rt2,
  title     = {{RT-2}: Vision-Language-Action Models Transfer Web
               Knowledge to Robotic Control},
  author    = {Brohan, Anthony and Brown, Noah and Carbajal,               Justice and others},
  booktitle = {Conference on Robot Learning},
  publisher = {PMLR},
  year      = {2023}
}

@inproceedings{ghosh2024octo,
  title     = {Octo: An Open-Source Generalist Robot Policy},
  author    = {Ghosh, Dibya and Walke, Homer Rich and                      Pertsch, Karl and others},
  booktitle = {Robotics: Science and Systems},
  year      = {2024},
}

@inproceedings{kim2025openvla,
  title     = {{OpenVLA}: An Open-Source Vision-Language-Action Model},
  author    = {Kim, Moo Jin and Pertsch, Karl and Karamcheti,              Siddharth and others},
  booktitle = {Conference on Robot Learning},
  publisher = {PMLR},
  year      = {2025}
}

@inproceedings{yang2026actor,
  title={Actor-critic for continuous action chunks: A reinforcement learning framework for long-horizon robotic manipulation with sparse reward},
  author={Yang, Jiarui and Zhu, Bin and Chen, Jingjing and others},
  booktitle={AAAI Conference on Artificial Intelligence},
  year={2026}
}

@inproceedings{nakamoto2025steering,
  title={Steering Your Generalists: Improving Robotic Foundation Models via Value Guidance},
  author={Nakamoto, Mitsuhiko and Mees, Oier and Kumar, Aviral and others},
  booktitle={Conference on Robot Learning},
  publisher={PMLR},
  year={2025}
}

@inproceedings{mirchandani2025scale,
  title     = {So You Think You Can Scale Up Autonomous Robot
               Data Collection?},
  author    = {Mirchandani, Suvir and Belkhale, Suneel and
               Hejna, Joey and others},
  booktitle = {Conference on Robot Learning},
  publisher = {PMLR},
  year      = {2025}
}

@inproceedings{black2025pi0,
  title     = {{$\pi_0$}: A Vision-Language-Action Flow Model for
               General Robot Control},
  author    = {Black, Kevin and Brown, Noah and Driess, Danny              and others},
  booktitle = {Proceedings of Robotics: Science and Systems},
  year      = {2025},
}

@article{lei2025rl100,
  title   = {Performant Robotic Manipulation with Real-World
             Reinforcement Learning},
  author  = {Lei, Kun and Li, Huanyu and Yu, Dongjie and others},
  journal = {Science Robotics},
  volume  = {11},
  number  = {116},
  pages   = {ead6267},
  year    = {2026},
  doi     = {10.1126/scirobotics.aed6267}
}

@article{wang2026learning,
  title         = {Learning While Deploying: Fleet-Scale Reinforcement
                   Learning for Generalist Robot Policies},
  author        = {Wang, Yi and Li, Xinchen and Xie, Pengwei and
                  others},
  journal       = {arXiv preprint arXiv:2605.00416},
  year          = {2026},
  eprint        = {2605.00416},
  archivePrefix = {arXiv},
}

\end{document}